# Multi-Scale DenseNet-Based Electricity Theft Detection

Bo Li[1], Kele Xu[2, 3], Xiaoyan Cui[1, *], Yiheng Wang[4], Xinbo Ai[1], Yanbo Wang[5]

1. Beijing University of Posts and Telecommunications,
Beijing, 100876, China
deepblue.lb@gmail.com
2. School of Computer, National University of Defense Technology,
Changsha, 410073, China
kelele.xu@gmail.com
3. School of Information Communication, National University of Defense Technology,
Wuhan, 430015, China
4. The University of Melbourne, Parkville, 3010, Australia
yihengw1@student.unimelb.edu.au
5. China Minsheng Bank,
Beijing 100031, China
wangyanbo@cmbc.com.cn

Electricity theft detection issue has drawn lots of
attention during last decades. Timely identification of
the electricity theft in the power system is crucial for
the safety and availability of the system. Although
sustainable efforts have been made, the detection task
remains challenging and falls short of accuracy and
efficiency, especially with the increase of the data
size. Recently, convolutional neural network-based
methods have achieved better performance in comparison
with traditional methods, which employ handcrafted
features and shallow-architecture classifiers. In this
paper, we present a novel approach for automatic
detection by using a multi-scale dense connected
convolution neural network (multi-scale DenseNet) in
order to capture the long-term and short-term periodic
features within the sequential data. We compare the
proposed approaches with the classical algorithms, and
the experimental results demonstrate that the multi-
scale DenseNet approach can significantly improve the
accuracy of the detection. Moreover, our method is
scalable, enabling larger data processing while no
handcrafted feature engineering is needed.

---

* Corresponding author

# 1 Introduction

In many countries, power utilities suffered a dramatic economic loss due to the electricity theft behavior, which has been a notorious problem in traditional power systems. Take the United States as an example, electricity theft makes power suppliers lose around six billion dollars every year [1]. In China, the electricity theft phenomenon is more common than before. Compared to previous mechanical meter, smart meter is more vulnerable as it turns to suffer more advanced attacks. Thus, timely identification of the electricity theft in the power system is crucial for its safety and availability.

There are two kinds of losses, which affect the management and the safety of power utilities [2]: 1) Technical loss caused by internal actions in the power system. They are mainly induced from electrical system components such as transmission lines and power transformers; 2) Non-Technical Losses (NTL), which were usually caused by external operation in the power system or electricity theft behaviors.

As NTL has been a major problem for most of energy companies, in this paper, we focus on the NTL detection leveraging the data of daily electricity consumption, with the aim to improve the accuracy of the electricity theft detection. Currently, NTL electricity theft detection methods can be divided into three groups:

1. State-based detection [3, 4]: by using improved sensors and other special devices to monitor user's behavior, the advanced equipment can improve the accuracy of electricity theft detection and reduce the false-positive rate. However, this method requires lots of equipment costs and software costs;
2. Game theory-based detection [5], a game between the electric utility and the thief is created. Although these methods cannot lead to an optimal solution, a low-cost and reasonable result may be achieved to relief energy theft. However, it is still a challenge to formulate the utility function of all players. Here the players include distributors, regulators and thieves. In addition, formulating potential strategies is also an issue;
3. Machine learning-based detection [6, 7, 8, 9, 10], by employing the machine learning algorithms to learn the user's electricity consumption behavior data achieved from sensors, a classifier can be trained to detect fraudulent users and recognize their behavior patterns. As the power enterprises have mass of user's historical electricity consumption data, this technique can make full use of them. The data mining strategy cannot only enhance the accuracy of detection to reduce the error rate, but also economize high software costs and equipment costs, avoid faults due to human intervention.

The machine learning-based approach has gained great interests as the detection accuracy has been improved consistently by using larger data. On the other hand, this kind of approach is privacy-preserving as only daily consumption is used to explore the theft pattern. Indeed, sustainable efforts have been made using different machine-learning approaches. However, all these classifiers relied on the handcrafted features, which are domain-specific and data-specific. For example, different features provide different performance, and the performance varies dramatically on different data by using the same feature sets.

Recently, deep learning methods have achieved dramatic success in different fields, such as, image classification [11], speech recognition [12]. Unlike previous efforts, which employ the handcrafted features and the shallow-architecture classifier, deep learning can be used to learn the feature in an automatic manner without the domain-specific constraints. On the other hand, with the dramatic increase of the data size, how to make fully use the big-data draws lots of attention, especially with the advance of the deep learning technique in the machine-learning field.

Here, we explore the use of convolutional neural network [11] for the detection task. Inspired by the statistical model approach, the periodicity of sequential data is of great importance for the classifier, and the series may have weekly, monthly, seasonal or annual periodicity. For the electricity theft detection, the pattern of electricity consumption is very salient for different user. Thus, efficient description the periodicity can be very helpful to improve the accuracy of electricity theft detection. Concretely, we propose to modify the multi-scale DenseNet, which can automatically capture the long-term and short-term periodic features of the sequential data. In summary, our contributions come with two folds:

- We propose to use DenseNet-based convolutional neural network to analyze the time series, with the aim to predict the probability of the theft behavior. Electricity consumption data has very strong periodicity, and convolutional neural networks can automatically capture some periodic features, different structures of networks can capture different kinds of periodic features, thus overcome the problems that classical machine learning model cannot get the periodic features;
- To capture more salient and abstract periodic features, we propose to modify the traditional dense block. Since the multi-scale dense block is given, the periodic features can be extracted within different time scales. Moreover, we suggest modifying the connection sequence of the layers in the dense block. On one hand, it greatly reduces the vibration of the training error, which makes the prediction more stable. On the other hand, it improves the accuracy significantly and provides superior performance.

Based on the aforementioned modifications, the accuracy of detection has been significantly improved. The rest of the paper is organized as follows. Section 2 gives the short summary on related work. Section 3 presents the proposed methods, while Section 4 provides the comparison experiment results using various algorithms. In the end, Section 5 draws conclusions of this paper.

## 2    Related Work

Sustainable efforts have been made to analyze the user's behavior patterns of electricity consumption, using the machine learning algorithm to establish a classification model to solve the problem. In summary, the classification methods include: statistical models [13], random forest [14], neural network [15], Support Vector Machines (SVM) [16]. In more detail, in [7], the author used SVM to detect electricity theft based on user's historical electricity consumption data over two years. This method used the past electricity consumption data. The problem is the longer-term data may be a noise for the model. In addition, it is of difficulties for the SVM classifier to learn the periodic features directly from the raw data. Moreover, the SVM is very time-consuming when using a large amount of data.

In [10], the author made an attempt to solve the problem by combining the decision tree and the SVM. This strategy improved the accuracy compared with the single SVM. However, this method is still hard to describe the periodic features. In [6], the author made use of statistical features, including the mean of power consumption, minimum of consumption, maximum of consumption, sum of consumption, standard deviation of power consumption, etc. Then, the k-means algorithm was employed to detect abnormal behavior. This kind of statistical features only depicts user's electricity consumption patterns in a certain aspect, which is not enough to improve the detection accuracy greatly. The handcrafted features replied on the domain knowledge, and the dimension of training data would be very large, which would induce a sharp increase in training time, Moreover, the features are task-dependent and data-dependent.

In [9], the author proposed to employ an ensemble algorithm to improve the accuracy of the model. Various machine learning algorithms are combined, including decision tree, SVM and optimum path forest, which improve the accuracy of the model by 2%-10%. However, the computing cost is dramatically increased. This is not applicable to the real case in the practical industrial world.

## 3    DenseNet-based electricity theft detection

In this section, we firstly, explore the traditional CNN to classify the electricity daily consumption time-series, then the DenseNet-based classification method is given. Moreover, the proposed multi-scale DenseNet is described. To make a quantitative comparison between CNN-based methods [17, 18, 19, 20]and the traditional hand-crafted features-based shallow architecture classifiers, random forest and gradient boosting machines are also tested in our experiments, which can act as the benchmarks.  And a short summary of the random forest and gradient boosting machines is given at the end of this part.

### 3.1 Convolutional Neural Network

The main difference between CNNs and traditional neural networks is that: CNNs have an automatic feature extractor, which consists of a convolution layer and a down-sampling layer (or pooling layer). Generally, a convolution layer includes a couple of feature maps, and each has some neurons. Normally, the parameters of the convolution kernel are initialized randomly (or using same specified initialization methods), and will be adjusted during the training step. The advantages achieved by the convolution kernel are that: the connections between each layer are reduced, and it decreases the risk of overfitting. Sub-sampling is also called pooling, and it usually includes mean sampling and maximum sampling methods [22]. The convolution layer and the sub-sampling layer lower the complexity of the model and substantially reduce the number of parameters [24]. In this paper, as our task can be viewed univariable time-series classification, 1D-CNN is employed in this paper.

### 3.2 DenseNet

DenseNet is a CNN architecture with dense connections, which provides dramatic performance improvement with comparison to previous CNN architectures. The DenseNet consists of many dense blocks, and each dense block connects to a transition layer except the last one which connects to the global pooling and the fully-connected layer directly. There is a direct connection between any two layers in each dense block. The input to each layer of the network is the union of the outputs from all the previous layers. In the network, each layer is connected with the previous layer directly in order to re-utilize features. At the same time, each layer is designed narrow to restrict the number of feature maps that can be learned and reduce the redundancy. Compared with the classical CNN, DenseNet not only performs better in image classification, but also has a higher utilization rate for the original data and less feature information loss. It reinforces feature propagation, supports feature re-utilization, solves vanishing gradient problems effectively and significantly reduce the number of parameters to economize the training time.

In the architecture, the transition Layer is a stacked series of nonlinear transformations, including Bach Normalization, ReLU and 1D-convolutional layer. Notably, the last Dense Block connects to a average pooling layer and a fully connected layer instead of a transition layer. Finally, the classification result is obtained by sigmoid function.

### 3.3 Multi-scale DenseNet

In the classical DenseNet, the size of the convolution kernel is very small (such as 3), thus it is difficult to get the middle-term and long-term periodical features, which are very important to improve the classification accuracy. Moreover, the kernel size is fixed in a small range. Different kinds of periodical features will be achieved if use a large range, and will help to get more abstract features after combine all the periodical features. Finally, if only modify the classical DenseNet in order to fit the one-dimensional data rather than change the whole structure, the outputs of the model will be unstable. Therefore, the traditional DenseNet architecture is not suitable for the electricity theft detection. To address aforementioned issues, two main modifications are proposed for the DenseNet architecture:

- The new multi-scale dense block is designed with a goal to capture the long-term and short-term periodic features within the sequential data. There are 24 different convolution layers in each multi-scale Dense Block, each layer has different length of convolution kernels. The length of the convolution kernel is longer in the upper layer and can generate more feature maps to extract long-term and medium-term user electricity features. In the lower layer, the convolution kernel receives all outputs in the previous layers and the original input. In consequence, it allows us to decrease the length of the convolution kernel and prompt the feature maps to grasp more short-term periodic features of electricity consumption. Since the network structure supports to re-utilize the feature maps, the multi-scale dense block concatenates all periodic features with different lengths into the transition layer for further manipulations. After this block, the model can extract a certain number of long and short-term user electricity behavior features from the original features and integrate them into the next layer.
- According to the Bach normalization-ReLU-Convolution layer connections in classical dense blocks, the outputs of 1D-DenseNet is unstable. In order to maintain the stability of the model and reduce the fluctuation of the outputs, the order of the connections between each convolution part in dense blocks is modified into Convolution-Bach normalization-ReLU (as shown in Fig. 1.)

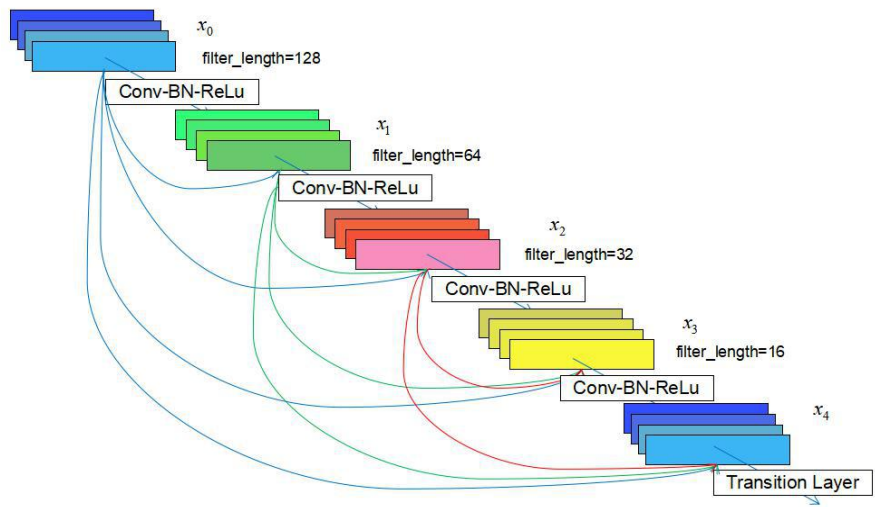

Fig. 1. Multi-scale dense block

### 3.4 Shallow Architecture Classifiers

In this part, a brief summary is conducted on two kinds widely-used shallow classifiers: Random Forest and Gradient Boosting Machines.

**Random Forest:**

Random Forest (RF) is a decision tree-based ensemble algorithm . By constructing multiple fully-grown decision trees, the final forecast is obtained by voting for each of their single prediction. In a RF model, all decision trees are trained independently, and usually do not need to prune. Different decision trees may learn different rules from the data. If it is a classification problem, each decision tree will get a prediction according to the result of its learning, then all the decision trees will vote to get the final prediction. If it is a regression problem, the prediction of each tree is averaged to get the final prediction. The stochastic characteristics for RF can be summarized in two aspects: sampling by rows and sampling by columns (features). These two strategies cannot only offset some negative effects induced by outliers, but also prevent overfitting to those significant features. To sum up, RF performs well in accuracy and has strong robustness.

**Gradient Boosting Machines:**

Gradient Boosting Machines (GBMs) [21] is a tree-based gradient boosting algorithm. By continuously fitting the residuals of the training samples, each new tree reduces the errors produced from the prediction of the previous tree. The strategy of reducing residuals greatly improves the prediction accuracy of the model. GBMs trains faster and more efficiently than many other algorithms, thus it is a popular machine learning algorithm.

## 4  Experimental Results

The data includes real electricity usage records of nearly 10,000 users collected in China. Each sample includes the daily electricity consumption data of the user over the past year. The data label is whether electricity theft behavior exits in the time-series. As we all know, the real data is often having missing values, we fill the missing values by using the Lagrange interpolation method [23]. Before feed the data into the model, Z-score standardization is employed to pre-process data. After using Z-score normalization, the data is transformed into the range which has 0 for mean and 1 for variance. On the one hand, standardization can reduce the impact of outliers on the model, and improve the performance, on the other hand, it can reduce the computation cost.

### 4.1 Handcrafted features

In accordance with users' historical electricity consumption data, a series of handcrafted features are created for the shallow-architecture classifiers. In brief, the features can be divided into three parts: 1) the maximum, minimum, mean, variance, mean and median for the records during the recent one month, two months, three months, six months and a year; 2) the average electricity consumption for each month; 3) the maximum, minimum, variance, median, skew and divergence for the data generated in part two. These three parts of features will be used in Random Forest and Gradient Boosting Machines.

### 4.2 Performance Measures

Logloss and AUC are used to evaluate the performance of different models [25, 26]. Logloss is also called cross entropy, which is a commonly used loss function for the classification. A lower logloss indicates better model performance.

AUC is a measure of the quality of the model. The AUC is the area of the area covered by the ROC curve. The value of AUC is generally between 0.5 and 1. Obviously, a higher AUC value corresponds to better performances of the model.

### 4.3 Experimental results

And the experimental whole flow chart is shown below. To ensure the stability and reliability of the results, all our experiments are conducted using 5 folds cross-validation. The whole framework of our experiments is given in Fig 2.

**Experiment 1: Random Forest:**

Firstly, the Random Forest method is tested on the data by using the aforementioned handcrafted features. By a large number of experiments, the ranges of parameters are shrunken within reasonable bounds. To sum up, the best prediction results for Random Forest are 0.289135 for logloss and 0.8430 for AUC.

**Experiment 2: Gradient Boosting Machines:**

LightGBM model, which is an excellent implementation of GBM, are employed in our experiments. In order to find the best performance by using GBM model, grid search is used to find the best parameters with small learning rate. The best performance using LightGBM is 0.287355 for logloss, and AUC is 0.8464.

**Experiment 3: Classical Convolutional Neural Network:**

An user's daily electricity consumption information for a year can be considered as a time series and manipulated by 1D convolution. The best results of classical CNN is 0.2833 for logloss, and AUC is 0.85728. In accordance with these experiments, when the length of the convolution kernel is 7, the model learns some weekly electricity consumption features. Whereas, when the value increases to 14 or even higher, the model cannot catch longer periodic features as we expected. Besides, two fully connection layers with 64 and 32 kernels is good enough to get a better result than LightGBM, both increase and decrease the number of kernels will reduce the performance of the model.

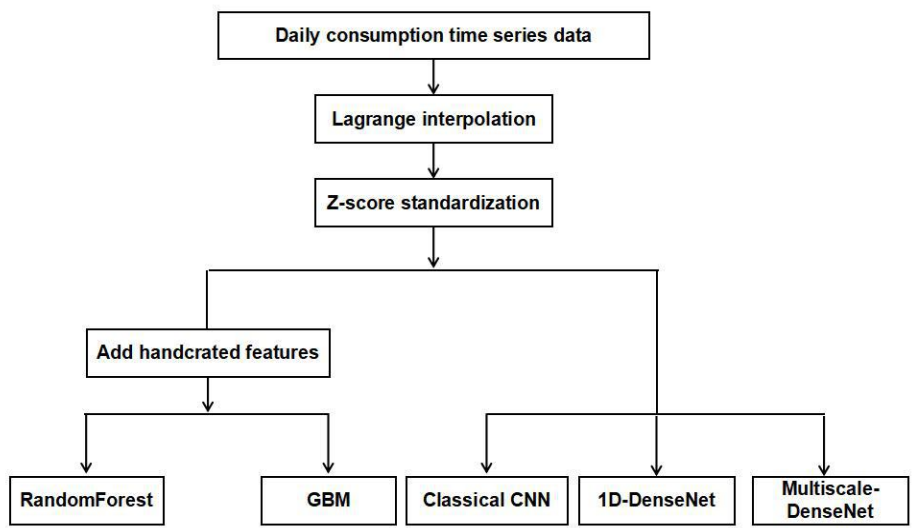

Fig. .2. Framework of Electricity Theft Detection Algorithms

**Experiment 4:1D-DenseNet**

After modified the convolution kernel in DenseNet into 1D-convolution kernel and properly changed the structure of the network, the improved 1D-DenseNet was employed into the dataset and achieved better results. The best performance using 1D-DenseNet is 0.27880 for logloss, and AUC is 0.86314. Take one 1D-DenseNet for example, there are two dense blocks in this structure and the first dense block is composed of 6 identical parts. There are two convolution layers in each part. The first layer has 128 kernels and the length of the kernel is 1. The second layer has 32 kernels and the length of the kernel is 3. Setting the length of the kernel 3 to get short-term periodical features, compared with 1D-DenseNet-A, the deeper structure is employed to extract more abstract features. However, if the model has 3 Dense Blocks, it tends to overfitting and does not perform well on the test set.

**Experiment 5: Multi-scale DenseNet**

The proposed Multi-scale DenseNet is based on the two aspects of improvement on traditional dense block. The first dense block is composed of 6 parts with same structures, each part has 4 convolution layers and their length of kernels decrease in order. The output of a layer will be merged into next layer's input and generate new features. This kind of structure extracts features in different periods and completely obtains periodic rules of electricity behaviors. As we can see from these structures, the network that contains a single multi-scale dense block performs very bad. The reason is that the model only extract various periodic features. The lack of those advanced characteristics leads to underfitting. By adding the second multi-scale dense block, the network boosts a lot due to the further abstraction for the caught features. we can learn that the performance has greatly been improved, and the outputs are more stable than the 1D-DenseNet. In all the models, Multi-scale DenseNet performances best. The improved logloss is 0. 2585 while the AUC value is 0. 8670.

## 5    Conclusion

This paper presents a novel classification method for electricity theft detection using the multi-scale DenseNet. The proposed method builds a new deep architecture for the CNN to investigate the time-series data. The CNN architecture mainly employs the multi-scale dense block to capture the salient patterns of the time-series data at different time scales.

In our experiments, both RF and GBM do not perform well for our time-series classification task, and the handcrafted features are domain-specific. Compared to the shallow-architectures classifiers, the conventional convolutional neural networks and 1D-DenseNet provide better performance as they can extract periodic features from the raw data. The proposed multi-scale DenseNet remarkably enhances the model performance and enabling for timely identification problem. The key advantages of proposed mutli-scale DenseNet are: 1). Multi-scale DenseNet can capture periodic information from the electricity records, in an end-to-end manner without handcrafted feature engineering. 2). The multi-scale dense block extracts seasonal trends in different periods, hence extracted features have more discriminative power, 3). Feature extraction and classification are unified in one CNN model, and their performances are mutually enhanced. Through the experiments, we demonstrate the proposed multi-scale DenseNet outperforms other models, and the AUC is improved from 0.8430 to 0.8670. We, therefore, believe that the proposed method can serve as a competitive tool of feature learning and classification of our time-series classification problem.